\documentclass{article} 
\usepackage{iclr2026_conference,times}

\usepackage{hyperref}
\usepackage{url}
\usepackage{csquotes}
\usepackage{enumitem}
\usepackage{booktabs}
\usepackage{longtable}
\usepackage{amssymb}

\usepackage{pgf}
\usepackage{pgfplots}
\usepackage{pgfplotstable}

\usepackage{listofitems, tikz}
\usetikzlibrary{shapes.geometric}
\usetikzlibrary{positioning}
\usetikzlibrary{arrows}
\usetikzlibrary{decorations.pathmorphing}
\usetikzlibrary{decorations.markings}
\usetikzlibrary{decorations.pathreplacing}
\usetikzlibrary{calc}
\usetikzlibrary{shadows}
\usetikzlibrary{arrows.meta}
\usetikzlibrary{bending}
\usetikzlibrary{intersections}
\usetikzlibrary{fadings}
\usetikzlibrary{chains}
\tikzfading[name=fade out, inner color=transparent!0,
  outer color=transparent!100]
\usetikzlibrary{babel}

\usetikzlibrary{backgrounds}
\usetikzlibrary{fit}
\definecolor{couleurpolice}{cmyk}{0.67, 0.66, 0, 0.71}
\definecolor{DeepBlue}{rgb}{0.03, 0.25, 0.53}
\definecolor{DeepGreen}{rgb}{0.00, 0.5, 0.00}
\definecolor{RedPink}{rgb}{.9,0,.2}
\definecolor{Alizarin}{rgb}{0.82, 0.1, 0.26}
\definecolor{Amber}{rgb}{1.0, 0.75, 0.00}
\definecolor{Auburn}{rgb}{0.43, 0.21, 0.1}
\definecolor{Byzantine}{rgb}{0.74, 0.2, 0.64}
\definecolor{CanaryYellow}{rgb}{1.0, 0.94, 0.0}

\definecolor{tgb1}{rgb}{0.2666, 0.466, 0.666}
\definecolor{tgb2}{rgb}{0.4, 0.8, 0.9333}
\definecolor{tgb3}{rgb}{0.1333, 0.5333, 0.2}
\definecolor{tgb4}{rgb}{0.8, 0.733, 0.2666}
\definecolor{tgb5}{rgb}{0.9333, 0.4, 0.466}
\definecolor{tgb6}{rgb}{0.666, 0.2, 0.466}

\definecolor{redrag}{rgb}{0.97254902, 0.901960784, 0.929411765}
\definecolor{bleurag}{rgb}{0.941176471, 0.952941176, 0.9529411760}
\definecolor{yellowrag}{rgb}{0.984313725, 0.952941176, 0.905882353}
\definecolor{purplerag}{rgb}{0.956862745, 0.937254902, 0.984313725}

\definecolor{ForestGreen}{RGB}{34,139,34}

\definecolor{Green}{RGB}{76, 119, 59}

\usepackage{soul} 
\definecolor{lichen}{rgb}{.91,.95,0.83}
\sethlcolor{lichen} 
\DeclareRobustCommand{\hlcgreen}[1]{{\sethlcolor{lichen}\hl{#1}}}

\DeclareRobustCommand{\hlcbluefonce}[1]{{\sethlcolor{tgb1}\hl{#1}}}

\colorlet{Mycolor1}{tgb1!20}

\colorlet{Mycolor2}{tgb2!40}

\colorlet{Mycolor3}{Byzantine!40}

\colorlet{Mycolor4}{Auburn!40}

\colorlet{Mycolor5}{gray!40}

\DeclareRobustCommand{\hlcyellowfonce}[1]{{\sethlcolor{tgb4}\hl{#1}}}

\title{Benchmarking Large Language Models for Quebec Insurance: From Closed-Book to Retrieval-Augmented Generation}

\author{David Beauchemin \& Richard Khoury \\
Group for Research in Artificial Intelligence of Laval University (GRAIL)\\
Université Laval\\
Quebec City, Quebec, Canada \\
\texttt{\{david.beauchemin, richard.khoury\}@ift.ulaval.ca}\\
}

\iclrfinalcopy
\begin{document}

\maketitle

\begin{abstract}
The digitization of insurance distribution in the Canadian province of Quebec, accelerated by legislative changes such as Bill 141, has created a significant \enquote{advice gap}, leaving consumers to interpret complex financial contracts without professional guidance. 
While Large Language Models (LLMs) offer a scalable solution for automated advisory services, their deployment in high-stakes domains hinges on strict legal accuracy and trustworthiness. 
In this paper, we address this challenge by introducing AEPC-QA, a private gold-standard benchmark of 807 multiple-choice questions derived from official regulatory certification (paper) handbooks. 
We conduct a comprehensive evaluation of 51 LLMs across two paradigms: closed-book generation and retrieval-augmented generation (RAG) using a specialized corpus of Quebec insurance documents. 
Our results reveal three critical insights: 
1) the supremacy of inference-time reasoning, where models leveraging chain-of-thought processing (e.g. \texttt{o3-2025-04-16}, \texttt{o1-2024-12-17}) significantly outperform standard instruction-tuned models; 
2) RAG acts as a knowledge equalizer, boosting the accuracy of models with weak parametric knowledge by over 35 percentage points, yet paradoxically causing \enquote{context distraction} in others, leading to catastrophic performance regressions; and 
3) a \enquote{specialization paradox}, where massive generalist models consistently outperform smaller, domain-specific French fine-tuned ones. 
These findings suggest that while current architectures approach expert-level proficiency ($\sim$79\%), the instability introduced by external context retrieval necessitates rigorous robustness calibration before autonomous deployment is viable.
\end{abstract}

\section{Introduction}
The financial sector is undergoing a profound transformation driven by digital distribution channels. 
In jurisdictions like Quebec (Canada), the implementation of Bill 141 allows insurers to distribute products entirely online, effectively removing the requirement of involving a human agent \citep{bill141}. 
While this legislative shift fosters market efficiency, it fundamentally transfers the burden of comprehension onto consumers, who are often ill-equipped to navigate insurance contracts exceeding 30,000 words and 50 pages \citep{beauchemin-etal-2020-generating, protectionepargnants, martinez2022poor, duclos2024droit}. 
It creates a \enquote{systemic vulnerability} characterized by significant information asymmetry and a lack of professional guidance---leaving laypeople to interpret \enquote{contradictory} endorsements that modify base coverage without assistance---a phenomenon scholars identify as the \enquote{advice gap} \citep{protectionepargnants}.

LLMs offer a potential solution for automated advisory services. 
However, in the insurance domain, \enquote{plausible} answers are insufficient; responses must be legally accurate. 
Misinformation can lead to coverage gaps and severe financial consequences for policyholders and legal action against insurers, as illustrated by recent litigations involving hallucinating chatbots \citep{air_canada}\footnote{In \citet{air_Canada_court}, the Civil Resolution Tribunal found the airline liable for \enquote{negligent misrepresentation} after its chatbot provided incorrect information regarding bereavement fares. The ruling established that the company was responsible for the accuracy of its automated agents, regardless of whether the error was human- or AI-driven.}.
Therefore, the deployment of LLMs in finance hinges on trustworthiness and factual consistency \citep{ressel2024addressing}.

This paper addresses the challenge of building a reliable insurance Question-Answering (QA) system by conducting an exhaustive benchmark of LLMs.
We systematically benchmark the intrinsic domain knowledge of these models in a closed-book setting \citep{roberts2020much} and evaluate the performance gains achieved through non-parametric context injection via RAG \citep{li2022survey}.
Our specific contributions are:
\begin{enumerate}[leftmargin=*, noitemsep, topsep=0ex]
    \item The introduction of AEPC-QA, a private gold-standard benchmark of 807 questions derived from regulator certification (print-only) handbooks \citep{amfbooks}, ensuring strict alignment with industry standards with low risk of data contamination.
    \item A comprehensive benchmark of 51 LLMs, comparing proprietary versus open-source models in a zero-shot setting.
    \item An empirical analysis demonstrating the impact of RAG architectures as a performance amplifier compared to the closed-book baseline for this specialized task.
\end{enumerate}

\section{Related Work}

\subsection{Legal and Insurance NLP Resources}
The application of NLP to the legal domain has accelerated with the release of large-scale benchmarks such as LegalBench \citep{guha2023legalbench} and LawBench \citep{fei2024lawbench}.
However, a significant \enquote{jurisdictional bias} persists: the vast majority of resources focus on \textbf{Common Law} systems (e.g. the US and the UK), where law is derived from judicial precedent \citep{gu2024survey}.
In contrast, \textbf{Civil Law} jurisdictions like Quebec rely on comprehensive codified statutes (the \enquote{Civil Code}) where interpretation requires distinct reasoning patterns.
This misalignment renders most \enquote{legal} LLMs (e.g. \texttt{SaulLM}, \texttt{LawGPT}) unsuitable for our context, as they are pre-trained on priors that do not exist in Quebec law.
While \citet{beauchemin2023risc} introduced the first synthetic contract corpus for Quebec, no gold-standard QA benchmark existed to rigorously test regulatory comprehension.

\subsection{Retrieval-Augmented Generation in Specialized Domains}
RAG has emerged as a dominant paradigm for knowledge-intensive tasks, particularly in law, where hallucinations are unacceptable \citep{li2022survey}. 
\citet{louis2024interpretable} demonstrated the efficacy of RAG for statutory law question answering in French, emphasizing the need to retrieve verifiable legal articles. 
Similarly, \citet{nuruzzaman2020intellibot} and \citet{beauchemin2024quebec} explored chatbot architectures for the insurance industry in two legal jurisdictions. 
However, current approaches often rely on general web-scraped knowledge bases, which may contain information that is outdated, incorrect, or from the wrong jurisdiction \citep{ressel2024addressing}. 
Our work addresses this by grounding generation in a curated, closed-domain corpus of official regulatory documents.

\subsection{Trustworthiness and Hallucination Benchmarking}
Evaluating factual consistency in financial advice is notoriously difficult.
Standard n-gram metrics (e.g. ROUGE, BLEU) have been shown to correlate poorly with human judgment of legal correctness, as they fail to capture semantic nuances, such as cases where a single negated word can reverse liability \citep{sulem2018bleu}.
While new benchmarks like FinTrust \citep{hu2025fintrust} attempt to assess safety and fairness, they are broad in scope and lack granular focus on specific insurance regulation.
Our study aligns with the findings of \citet{ho2024hallucinating}, which argue that \enquote{general} legal reasoning capability is a myth; models must be evaluated against the specific, granular statutes of the jurisdiction in which they will operate.

\section{Experimental Setup}
\subsection{Evaluation}
We evaluated various models across two distinct architectural paradigms to assess their capability in answering specialized insurance questions, namely 1) closed-book and 2) RAG-based approach. Closed-book evaluation means that we benchmark all models to assess their intrinsic knowledge of Quebec insurance law \citep{roberts2020much}. Next, to assess the impact of non-parametric knowledge injection, we implement an advanced RAG pipeline following the architecture proposed by \citet{beauchemin2024quebec}.
The system utilizes a dense retriever (i.e. text-embedding-ada-002) to index the knowledge corpus.
We prioritized a dense retrieval approach over sparse methods (such as BM25) because layperson queries often lack the precise legal terminology found in regulatory documents. Dense embeddings enable semantic matching between the user's intent and legal concepts, mitigating the vocabulary mismatch that often causes keyword-based retrievers to fail in specialized domains.
As our knowledge base, we use the \enquote{Quebec Automobile Insurance Expertise Reference Corpus} (QAIERC).
This corpus aggregates 168,600 sentences (2.6M tokens) of official insurance-related legislation, standardized contracts based on the regulator template, and regulatory guidance documents.
Upon query, the system retrieves the top-5 relevant chunks, which are then processed by a context compressor to reduce noise before augmenting the LLM's prompt.
The system prompt is presented in the \hlcbluefonce{blue} box in \autoref{fig:qnaprompt_exemple}.

\subsubsection{Evaluation Benchmark}
Evaluating generation quality in specialized domains is notoriously difficult \citep{cao2025toward}. 
To enable scalable and objective assessment in the Quebec insurance domain, we created the AMF Exam Preparation Content - QA (AEPC-QA) corpus.
We digitized (using Microsoft Optical Character Recognition \citep{azureaidoc}) and manually curated 25 official exam-preparation print handbooks used to certify insurance representatives.
The resulting dataset contains 807 multiple-choice questions, each with 4 choices and a single, objectively correct answer accepted by the regulator.
We present in \autoref{fig:qnaprompt_exemple} a translated instance of the corpus in the \hlcyellowfonce{yellow} box.
Due to copyright restrictions protecting the source material, this corpus remains private and cannot be publicly released.
Crucially, because these handbooks are available exclusively in paper format and are not publicly indexed online, there is minimal risk of data contamination from this corpus in pre-trained LLMs, ensuring a rigorous evaluation of unseen domain knowledge.

\begin{figure}[ht!]
    \centering
        \begin{tikzpicture}[scale=1, every node/.style={transform shape}]
            \node[rectangle, rounded corners, draw=tgb1, fill=tgb1, text width=0.99\linewidth, align=left, inner sep=1.2ex] (prompt) {$\ll$system/user$\gg$You're a damage insurance agent in Quebec. Your task is to answer multiple-choice questions about damage insurance in Quebec by selecting the correct answer. Answer in French. Answer only with the letter of your answer choice.
            };
            \node[rectangle, rounded corners, draw=tgb4, fill=tgb4, below=0.1cm of prompt, text width=0.99\linewidth, align=left, inner sep=1.2ex] (input) {
            $\ll$user$\gg$The question is:\\
            Armand has always been a keen golfer. Last week, while playing with friends on his favourite golf course, he was the victim of an unfortunate incident: while walking to a hole, he was hit in the side of the head by a golf ball and suffered serious injuries. Which of the following defences would be irrelevant to the golf club in the event of Armand taking legal action against it? The answer choices are:\\
            a) Advise all along the course to be cautious in your movements and to pay attention to the actions of other players.\\
            b) Acceptance of the risks involved in playing golf.\\
            \textbf{c) The Good Samaritan.}\\
            d) Sharing responsibility.\\
            };
        \end{tikzpicture}
    \caption{Example of a translated APEC-QA question, choices and \textbf{response} along with the prompt used for the evaluation. \hlcbluefonce{Blue} box contains the task instructions. \hlcyellowfonce{Yellow} box contains the prefix for the model to continue. Texts in \enquote{\texttt{$\ll\gg$}} are role-tags to be fed to the model.}
    \label{fig:qnaprompt_exemple}
\end{figure}

To provide insight into the linguistic complexity and domain specificity of this private benchmark, we present comprehensive descriptive statistics in \autoref{tab:dataset_stats}. 
The corpus comprises over 96,000 tokens in question text alone, with an average sentence length of 18.95 words, reflecting the verbose nature of regulatory scenarios. 
The readability score (Flesch-Kincaid \citet{Kincaid1975DerivationON} (the French version)) of 64.55 indicates a moderate difficulty level, consistent with professional certification examinations.

\begin{table}[ht!]
    \centering
    \caption{Descriptive Statistics of the AEPC-QA Dataset. Token counts and sentence lengths are reported as averages or totals across the corpus.}
    \label{tab:dataset_stats}
    \begin{tabular}{lr}
        \toprule
        \textbf{Metric} & \textbf{Value} \\
        \midrule
        Total Questions & 807 \\
        Total Tokens (Questions) & 96,080 \\
        Total Tokens (Justifications) & 75,147 \\
        Average Sentence Length (Questions) & 18.95 words \\
        Average Sentence Length (Justifications) & 17.47 words \\
        Vocabulary Size (Questions) & 6,753 \\
        Readability Score (Flesch-Kincaid) & 64.55 \\
        \bottomrule
    \end{tabular}
\end{table}

\subsubsection{Evaluation Protocol and Metrics}
We employ a stratified 10-fold cross-validation protocol to minimize data partition bias and report a mean and one standard deviation over the 807 multiple-choice questions using the following seeds: $[42, \cdots,  51]$.
Performance is measured by accuracy on multiple-choice questions. 
This evaluation protocol is a proxy for assessing domain knowledge retention and reasoning capabilities.

\subsection{Evaluated Models}
\subsubsection{Baselines}
To contextualize the LM's performance, we establish several baselines. 
These baselines represent naive or simplistic strategies and serve as a \enquote{floor} performance. 
All baselines are expected to obtain a score close to 25\% due to the expected statistical distribution of correct answer choices. 
They will provide a crucial reference point for interpreting the scores of more advanced systems we evaluate.
Undoubtedly, any sophisticated model should significantly outperform them to be considered effective. 
We use the following five baselines:

\begin{itemize}[leftmargin=*, noitemsep, topsep=0ex]
    \item \texttt{Random Sampling}: Approach that samples randomly one of the four choices using an initial random seed. We use the same seed as the 10-fold (i.e. $[42, 43, \cdots, 50, 51]$).
    \item \texttt{Always A/B/C/D}: Approach that leverage deterministic guessing, namely \textbf{always} sampling the choice \enquote{a}, \enquote{b}, \enquote{c}, or \enquote{d}.
\end{itemize}

\subsubsection{LLM}
To ensure a comprehensive evaluation of the current landscape, we evaluate 51 LLMs.
The full details of evaluated models are available in \autoref{an:selectedllmdetails}, and we present our hardware and budget in \autoref{an:hardware}.
Three primary criteria guided our selection process to maximize the relevance and robustness of the benchmark:

\begin{enumerate}[leftmargin=*, noitemsep, topsep=0ex]
    \item \textbf{Performance on General Benchmarks:} We prioritized models achieving top rankings on established leaderboards, specifically the HuggingFace Open LLM Leaderboard \citep{open-llm-leaderboard-v2} and the LMSYS Chatbot Arena \citep{zheng2023judging}. It ensures we are testing models proven to possess strong reasoning and instruction-following capabilities.
    \item \textbf{Architecture and Scale:} We selected models spanning a wide range of parameter counts (7B to 70B) and architectures (e.g. Mixture-of-Experts). It allows us to analyze the scaling laws of domain knowledge retention. The selection includes major families such as \texttt{Llama-3} and \texttt{Qwen-2.5}.
    \item \textbf{Proprietary vs. Open Source:} To establish a \enquote{performance upper bound,} we included leading proprietary models (e.g. \texttt{GPT-4o}) to compare against open-weight models, assessing the gap between open and closed weights.
\end{enumerate}

Notably, we explicitly excluded models fine-tuned on general legal or insurance corpora, such as \textit{SaulLM} \citep{colombo2024saullm} and \textit{LawGPT} \citep{nguyen2023brief}.
Current open-weighted legal models are predominantly trained on English-language data from Common Law jurisdictions (e.g. the United States or the United Kingdom) \citep{gu2024survey}.
Given that Quebec operates under a distinct Civil Law system and our task is strictly Francophone, these models possess jurisdictionally irrelevant priors that render them unsuitable for this specific use case, a limitation highlighted in recent cross-jurisdictional legal studies \citep{louis2024interpretable, magesh2024hallucination}.

\section{Results and Discussion}

The evaluation of 51 LLMs and 5 baselines across both closed-book and RAG settings reveals three distinct behavioural patterns that fundamentally alter our understanding of AI readiness for insurance advisory tasks.

\subsection{The Supremacy of Inference-Time Reasoning}
The most significant finding is the dominance of \enquote{reasoning} models—architectures that utilize chain-of-thought (CoT) processing at inference time.
As shown in \autoref{table:resqaiercrag}, \texttt{o3-2025-04-16} and \texttt{o1-2024-12-17} secure the top two positions, reaching accuracies of 78.68\% and 75.18\% respectively.
This superior performance, which outstrips standard instruction-tuned models like \texttt{GPT-4o} and \texttt{Claude 3.5 Sonnet}, suggests that the primary bottleneck in legal question answering is not merely knowledge retrieval, but the \textit{application} of that knowledge to novel scenarios.
The AEPC-QA benchmark, which presents complex liability scenarios, requires multi-step logic: identifying the relevant legal principle (e.g. acceptance of risk vs. civil liability) and applying it to the specific facts.
Reasoning models, by effectively \enquote{thinking} before generating, appear uniquely suited to navigate these logical dependencies, validating the hypothesis that legal advisory systems require distinct cognitive architectures compared to general-purpose chatbots.

Interestingly, preliminary analysis of the hidden CoT traces suggests that even for this Francophone benchmark, reasoning models often perform intermediate logical steps in English before generating the final output. 
It reinforces the hypothesis that the reasoning capabilities of current LLMs are cross-lingual transfers anchored in their English-dominant pre-training \citep{Barua2025}.

\subsection{Closed-Book versus RAG}
Our experiments strongly validate the hypothesis that RAG acts as a \enquote{knowledge equalizer}, particularly for models with weaker pre-training on Francophone legal data.
The most striking example is \texttt{DeepSeek-reasoner}. 
In the closed-book setting, it achieved a subpar accuracy of 36.30\%, indicating a lack of memorization of Quebec civil law.
However, when augmented with the QAIERC corpus and a RAG architecture, its performance surged to 71.77\%. 
These results demonstrate that high-quality reasoning engines do not need to memorize jurisdiction-specific statutes during pre-training; when they are supplied with the correct regulatory context, they can synthesize answers competitive with those of models that have seen more relevant training data, such as OpenAI LLMs and language-based LLMs.
This finding is pivotal for the insurance industry, as it suggests that smaller or less specialized models can be effectively adapted to niche insurance products through robust retrieval pipelines rather than expensive fine-tuning.

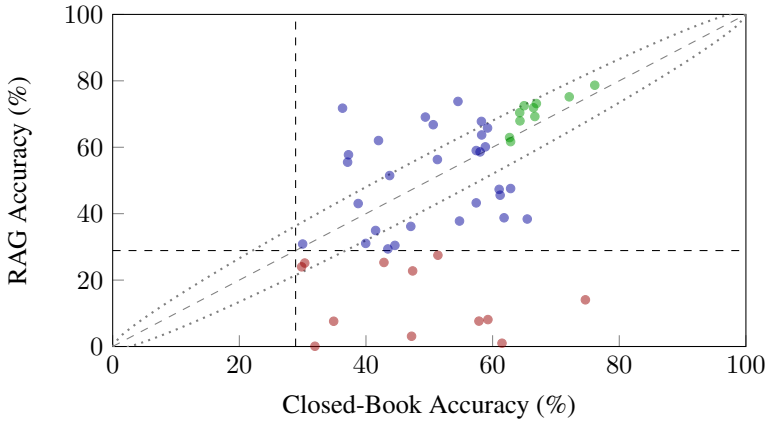
\begin{figure}[ht!]
    \centering
\begin{tikzpicture}
\begin{axis}[
    width=10cm, height=6cm,
    xlabel={Closed-Book Accuracy (\%)},
    ylabel={RAG Accuracy (\%)},
    xmin=0.00, xmax=100.0,
    ymin=0.00, ymax=100.0,
    xtick pos=left, 
    ytick pos=left, 
    grid=none,
    legend pos=north west,
    xticklabel style={/pgf/number format/fixed, /pgf/number format/precision=0}, 
    yticklabel style={/pgf/number format/fixed, /pgf/number format/precision=0}, 
    scaled ticks=false, 
]

\addplot[color=gray, dashed, domain=0:100] {x};

\draw[black, dashed] (axis cs:28.89, 0) -- (axis cs:28.89, 100);
\draw[black, dashed] (axis cs:0, 28.89) -- (axis cs:100, 28.89);

\addplot[color=gray, dotted, thick, smooth] coordinates {
    (0.00, 0.00) (1.01, 2.65) (2.02, 4.32) (3.03, 5.84) (4.04, 7.27) (5.05, 8.64) (6.06, 9.97) (7.07, 11.27) (8.08, 12.55) (9.09, 13.80) (10.10, 15.04) (11.11, 16.26) (12.12, 17.47) (13.13, 18.66) (14.14, 19.85) (15.15, 21.02) (16.16, 22.19) (17.17, 23.35) (18.18, 24.50) (19.19, 25.64) (20.20, 26.78) (21.21, 27.91) (22.22, 29.03) (23.23, 30.15) (24.24, 31.26) (25.25, 32.37) (26.26, 33.47) (27.27, 34.57) (28.28, 35.66) (29.29, 36.75) (30.30, 37.83) (31.31, 38.91) (32.32, 39.98) (33.33, 41.06) (34.34, 42.12) (35.35, 43.18) (36.36, 44.24) (37.37, 45.30) (38.38, 46.35) (39.39, 47.40) (40.40, 48.44) (41.41, 49.48) (42.42, 50.52) (43.43, 51.55) (44.44, 52.58) (45.45, 53.61) (46.46, 54.63) (47.47, 55.65) (48.48, 56.67) (49.49, 57.69) (50.51, 58.70) (51.52, 59.70) (52.53, 60.71) (53.54, 61.71) (54.55, 62.70) (55.56, 63.70) (56.57, 64.69) (57.58, 65.67) (58.59, 66.65) (59.60, 67.63) (60.61, 68.61) (61.62, 69.58) (62.63, 70.55) (63.64, 71.52) (64.65, 72.48) (65.66, 73.44) (66.67, 74.39) (67.68, 75.34) (68.69, 76.28) (69.70, 77.23) (70.71, 78.16) (71.72, 79.09) (72.73, 80.02) (73.74, 80.95) (74.75, 81.86) (75.76, 82.78) (76.77, 83.69) (77.78, 84.59) (78.79, 85.48) (79.80, 86.38) (80.81, 87.26) (81.82, 88.14) (82.83, 89.01) (83.84, 89.87) (84.85, 90.72) (85.86, 91.57) (86.87, 92.40) (87.88, 93.23) (88.89, 94.04) (89.90, 94.84) (90.91, 95.62) (91.92, 96.38) (92.93, 97.13) (93.94, 97.85) (94.95, 98.54) (95.96, 99.19) (96.97, 99.78) (97.98, 100.28) (98.99, 100.63) (100.00, 100.00)
};

\addplot[color=gray, dotted, thick, smooth] coordinates {
    (0.00, 0.00) (1.01, -0.63) (2.02, -0.28) (3.03, 0.22) (4.04, 0.81) (5.05, 1.46) (6.06, 2.15) (7.07, 2.87) (8.08, 3.62) (9.09, 4.38) (10.10, 5.16) (11.11, 5.96) (12.12, 6.77) (13.13, 7.60) (14.14, 8.43) (15.15, 9.28) (16.16, 10.13) (17.17, 10.99) (18.18, 11.86) (19.19, 12.74) (20.20, 13.62) (21.21, 14.52) (22.22, 15.41) (23.23, 16.31) (24.24, 17.22) (25.25, 18.14) (26.26, 19.05) (27.27, 19.98) (28.28, 20.91) (29.29, 21.84) (30.30, 22.77) (31.31, 23.72) (32.32, 24.66) (33.33, 25.61) (34.34, 26.56) (35.35, 27.52) (36.36, 28.48) (37.37, 29.45) (38.38, 30.42) (39.39, 31.39) (40.40, 32.37) (41.41, 33.35) (42.42, 34.33) (43.43, 35.31) (44.44, 36.30) (45.45, 37.30) (46.46, 38.29) (47.47, 39.29) (48.48, 40.30) (49.49, 41.30) (50.51, 42.31) (51.52, 43.33) (52.53, 44.35) (53.54, 45.37) (54.55, 46.39) (55.56, 47.42) (56.57, 48.45) (57.58, 49.48) (58.59, 50.52) (59.60, 51.56) (60.61, 52.60) (61.62, 53.65) (62.63, 54.70) (63.64, 55.76) (64.65, 56.82) (65.66, 57.88) (66.67, 58.94) (67.68, 60.02) (68.69, 61.09) (69.70, 62.17) (70.71, 63.25) (71.72, 64.34) (72.73, 65.43) (73.74, 66.53) (74.75, 67.63) (75.76, 68.74) (76.77, 69.85) (77.78, 70.97) (78.79, 72.09) (79.80, 73.22) (80.81, 74.36) (81.82, 75.50) (82.83, 76.65) (83.84, 77.81) (84.85, 78.98) (85.86, 80.15) (86.87, 81.34) (87.88, 82.53) (88.89, 83.74) (89.90, 84.96) (90.91, 86.20) (91.92, 87.45) (92.93, 88.73) (93.94, 90.03) (94.95, 91.36) (95.96, 92.73) (96.97, 94.16) (97.98, 95.68) (98.99, 97.35) (100.00, 100.00)
};

\addplot[only marks, mark=*, color=red!60!black, mark options={scale=0.8, fill=red!60!black}, opacity=0.5] coordinates {
    (51.36, 27.49) (42.84, 25.31) (30.33, 25.14) (29.84, 23.99) (47.37, 22.76) (74.65, 14.06) (59.26, 8.11) (57.84, 7.65) (34.90, 7.61) (47.20, 3.09) (61.48, 0.99) (31.94, 0.08)
};

\addplot[only marks, mark=*, color=green!60!black, mark options={scale=0.8, fill=green!60!black}, opacity=0.5] coordinates {
    (76.13, 78.68) (72.10, 75.18) (66.91, 73.17) (64.98, 72.51) (66.46, 71.89) (64.28, 70.41) (66.67, 69.26) (64.32, 67.94) (62.68, 62.92) (62.84, 61.69)
};

\addplot[only marks, mark=*, color=blue!60!black, mark options={scale=0.8, fill=blue!60!black}, opacity=0.5] coordinates {
    (54.53, 73.80) (36.30, 71.77) (49.38, 69.09) (58.23, 67.78) (50.62, 66.79) (59.20, 65.80) (58.27, 63.70) (41.98, 62.02) (58.85, 60.12) (57.41, 59.01) (58.02, 58.60) (37.24, 57.74) (51.28, 56.30) (37.10, 55.51) (43.74, 51.48) (62.84, 47.57) (61.03, 47.32) (61.19, 45.56) (57.41, 43.25) (38.77, 43.05) (61.81, 38.76) (65.47, 38.39) (54.77, 37.78) (47.08, 36.17) (41.52, 34.94) (39.96, 31.05) (30.00, 30.86) (44.57, 30.45) (43.46, 29.38)
};

\end{axis}
\end{tikzpicture}
    \caption{Accuracy scores with (y-axis) and without (x-axis) RAG system. \textcolor{black}{\textbf{Black}} dashed lines are our \texttt{Random} baseline scores. The dotted lines represent the statistical significance boundaries ($Z=\pm 3.29$, $\alpha=0.001$). Points falling within the central cone indicate no statistically significant difference between closed-book and RAG performance. Points above the upper dotted line show significant improvement with RAG, and points below mean the opposite. \textcolor{red!60!black}{\textbf{Red}} dots are models that performed poorer than the baseline on one of the corpora, \textcolor{green!60!black}{\textbf{green}} dots are models that performed better than 60\% on both corpora, while \textcolor{blue!60!black}{\textbf{blue}} dots are those that do not fit in the two other performance classes.}
    \label{fig:report}
\end{figure}

Moreover, since one of our objectives is to see whether using a different LLM can improve overall performance, we compare each LLM's results using a Z-test for statistical significance \cite{newcombe1998two}. 
Our null hypothesis is that the pair of scores are equal (i.e. no significant difference between the two results), meaning that values outside $|3.290527|$ allow us to reject the hypothesis with $\alpha = 0.001$. 
A positive value means that the first approach (with RAG) has a significantly better performance than the second (without RAG), and a negative value means the opposite. 
We report the aggregated results in \autoref{fig:report} with a visualization of the Z-test results, where the dash-dotted diagonal line represents equal performance across both tests, and the dotted lines represent the statistical significance bands.
A visual inspection reveals three sets of performance results. Systems that have been extensively trained or fine-tuned in this domain (in \textcolor{green!60!black}{\textbf{green}})) receive little benefit from the RAG architecture. Conversely, the cluster of \textcolor{red!60!black}{\textbf{red}} points highlights a subset of models that failed to outperform the random baseline ($28.89\%$), indicating that a RAG cannot compensate for models that lack fundamental closed-book or zero-shot reasoning capabilities. Intermediate models indicated in \textcolor{blue!60!black}{\textbf{blue}} are those that have not been trained or fine-tuned to have high closed-book performances. Most of these LLMs benefit from the RAG architecture, in some cases even surpassing the green models.


\subsection{The \enquote{Context Distraction} Phenomenon}
Conversely, the data reveals a vulnerability we term \enquote{context distraction}, in which the introduction of external context, in the form of a complex prompt and reference documents, causes a catastrophic performance regression.
It is well illustrated by \texttt{Gemini-2.5-pro}.
In the closed-book setting, this model exhibited strong intrinsic knowledge, scoring 74.65\% (placing it 2nd overall).
However, under the RAG condition, its accuracy collapsed to 14.06\%, a performance significantly worse than random guessing (25\%).
Similar collapses were observed in \texttt{Llama-3.3-70b-it} and \texttt{Grok-2-latest}.

We observe two primary causes for this failure, distinct from simple hallucinations. 
The first is format non-compliance due to verbosity. As noted in our analysis of failure cases, models like \texttt{Gemini-2.5-pro} often retrieved relevant information but failed to adhere to the strict single-letter output constraint, instead generating lengthy, nuanced explanations that could not be parsed as a valid choice, leading to a technical score of zero despite the reasoning potentially being correct.
The second cause is safety refusal triggers. Models with rigid safety alignment, such as \texttt{Llama-3.3}, frequently interpreted the complex RAG prompt context—which contains legal liability descriptions—as a sensitive query, triggering refusal mechanisms (e.g. 
\enquote{I cannot provide legal advice}) rather than attempting to answer the multiple-choice question.

This indicates that \enquote{context distraction} manifests differently depending on the model's safety alignment tuning.
This finding highlights a severe deployment risk: encapsulating an LLM in a RAG is not a guaranteed performance amplifier. 
Without rigorous model-specific calibration, a RAG can actively lobotomize highly capable models, rendering them unsafe for client-facing advisory roles.

\subsection{The Limits of Domain Specialization}
Our analysis regarding the fourth selection criterion—domain and language specialization—reveals a clear \enquote{specialization paradox}.
Contrary to the expectation that models specifically fine-tuned for the target language (French) or specific instruction-following tasks would achieve top results, they frequently underperformed compared to massive generalist models.

It is best exemplified by the performance of specialized variants like \texttt{French-Alpaca-Llama3-8b-it}, which ranks last on the leaderboard, achieving 0.08\% accuracy with the RAG architecture.
While such models adapt vocabulary and syntax to the target domain, they appear to sacrifice the \enquote{reasoning density} required to navigate the complex logical traps embedded in the evaluation corpus.
In contrast, generalist models like \texttt{o3-2025-04-16} and \texttt{Claude-3.5-Sonnet}, despite being trained on broad, multilingual corpora, exhibit superior zero-shot transfer.
This suggests that, for financial advisory tasks, the ability to reason abstractly about liability and causality (a general capability) is far more critical than surface-level alignment with the specific legal jargon or language of the jurisdiction.

\paragraph{The Economic Trade-off}
While reasoning models achieve superior accuracy, they impose a significant inference latency and cost premium.
For instance, \texttt{o3-2025-04-16} requires approximately 45 seconds of \enquote{thinking time} per query to achieve a 76.13\% accuracy.
In contrast, \texttt{GPT-4o-mini} with a RAG architecture achieves a respectable 65.80\% at a fraction of the compute cost.
For real-time customer-facing applications, this trade-off suggests a tiered deployment strategy: using lightweight RAG models for general inquiries and escalating to reasoning models only for complex liability assessments.

\subsection{The Proprietary Gap}
Finally, the benchmark highlights a persistent gap between proprietary and open-weight models in this specialized domain.
While open models like \texttt{Pixtral-large-latest} perform respectably (58.60\% with RAG), they lag approximately 20 percentage points behind the state-of-the-art proprietary models.
This suggests that for high-stakes/high-liability tasks such as insurance advice, the current generation of open-source models may not yet possess the requisite reasoning density or multilingual robustness to serve as standalone autonomous agents.

\begin{table}[ht!]
    \centering
    \caption{Comparative performance of selected LLM using a \enquote{closed-book} approach (left) or using a RAG architecture (right). 
    The performance is reported as the mean and one standard deviation.
    The table ranks models by the second-column accuracy ($\blacktriangledown$). 
    Models with \enquote{it} in their name are instructed models.
    Models below the double line are worse than the \texttt{Always C} baseline.
    We use \hlcgreen{colour} to emphasize our baselines.
    \underline{Underline} results are those that the LLM with an RAG architecture increase over the sole LLM. $\uparrow$ means higher is better.
    }.
    \label{table:resqaiercrag}
    \resizebox{\textwidth}{!}{%
    \begin{tabular}{clccc}
        \toprule
        Rank & LLM & \begin{tabular}[c]{@{}c@{}}Accuracy Closed-Book (\%) ($\uparrow$)\end{tabular} & \begin{tabular}[c]{@{}c@{}}Accuracy RAG (\%) ($\uparrow$) $(\blacktriangledown)$\end{tabular}\\
        \midrule
        1 & \texttt{o3-2025-04-16} & 76.13 $\pm$ 2.57 & \underline{78.68 $\pm$ 2.70} \\
        2 & \texttt{o1-2024-12-17} & 72.10 $\pm$ 2.35 & \underline{75.18 $\pm$ 1.89} \\
        3 & \texttt{Sonar-deep-research} & 54.53 $\pm$ 6.10 & \underline{73.80 $\pm$ 2.11} \\
        4 & \texttt{Claude-Opus-4-20250514} & 66.91 $\pm$ 3.00 & \underline{73.17 $\pm$ 2.13} \\
        5 & \texttt{Claude-3-7-Sonnet-20250219} & 64.98 $\pm$ 1.89 & \underline{72.51 $\pm$ 2.16} \\
        6 & \texttt{Claude-Sonnet-4-20250514} & 66.46 $\pm$ 2.52 & \underline{71.89 $\pm$ 1.87} \\
        7 & \texttt{DeepSeek-reasoner} & 36.30 $\pm$ 2.79 & \underline{71.77 $\pm$ 2.57} \\
        8 & \texttt{o4-mini-2025-04-16} & 64.28 $\pm$ 3.66 & \underline{\textit{70.41 $\pm$ 2.71}} \\
        9 & \texttt{GPT-$4.5$-preview-2025-02-27} & 66.67 $\pm$ 2.28 & \underline{\textit{69.26 $\pm$ 2.49}} \\
        10 & \texttt{R1-1776} & 49.38 $\pm$ 2.42 & \underline{\textit{69.09 $\pm$ 1.84}} \\
        11 & \texttt{GPT-$4.1$-2025-04-14} & 64.32 $\pm$ 2.67 & \underline{\textit{67.94 $\pm$ 1.92}} \\
        12 & \texttt{DeepSeek-chat} & 58.23 $\pm$ 2.67 & \underline{\textit{67.78 $\pm$ 2.38}} \\
        13 & \texttt{Sonar-reasoning-pro} & 50.62 $\pm$ 2.88 & \underline{\textit{66.79 $\pm$ 4.15}} \\
        14 & \texttt{GPT-$4.1$-mini-2025-04-14} & 59.20 $\pm$ 2.32 & \underline{65.80 $\pm$ 2.75} \\
        15 & \texttt{GPT-4-0613} & 58.27 $\pm$ 1.89 & \underline{63.70 $\pm$ 2.71} \\
        16 & \texttt{Grok-3-latest} & 62.68 $\pm$ 2.49 & \underline{62.92 $\pm$ 3.07} \\
        17 & \texttt{o1-mini-2024-09-12} & 41.98 $\pm$ 3.60 & \underline{62.02 $\pm$ 2.86} \\
        18 & \texttt{Grok-3-fast-latest} & 62.84 $\pm$ 2.88 & 61.69 $\pm$ 3.54 \\
        19 & \texttt{o3-mini-2025-01-31} & 58.85 $\pm$ 2.94 & \underline{60.12 $\pm$ 2.11} \\
        20 & \texttt{Claude-3-5-Haiku-20241022} & 57.41 $\pm$ 2.86 & \underline{59.01 $\pm$ 2.15} \\
        21 & \texttt{Pixtral-large-latest} & 58.02 $\pm$ 2.33 & \underline{58.60 $\pm$ 3.87} \\
        22 & \texttt{Qwen-max} & 37.24 $\pm$ 3.23 & \underline{57.74 $\pm$ 1.56} \\
        23 & \texttt{GPT-4o-mini-2024-07-18} & 51.28 $\pm$ 3.40 & \underline{56.30 $\pm$ 1.93} \\
        24 & \texttt{Phi-4} & 37.10 $\pm$ 15.10 & \underline{55.51 $\pm$ 3.95} \\
        25 & \texttt{GPT-$4.1$-nano-2025-04-14} & 43.74 $\pm$ 4.79 & \underline{51.48 $\pm$ 2.14} \\
        26 & \texttt{Mistral-medium-2505} & 62.84 $\pm$ 1.78 & 47.57 $\pm$ 2.55 \\
        27 & \texttt{Grok-3-mini-fast-latest} & 61.03 $\pm$ 4.17 & 47.32 $\pm$ 4.04 \\
        28 & \texttt{Grok-3-mini-latest} & 61.19 $\pm$ 3.15 & 45.56 $\pm$ 2.07 \\
        29 & \texttt{Mistral-large-latest} & 57.41 $\pm$ 2.19 & 43.25 $\pm$ 3.15 \\
        30 & \texttt{Sonar} & 38.77 $\pm$ 1.57 & \underline{43.05 $\pm$ 3.61} \\
        31 & \texttt{Magistral-medium-2506} & 61.81 $\pm$ 2.93 & 38.76 $\pm$ 2.84 \\
        32 & \texttt{Gemini-$2.5$-flash} & 65.47 $\pm$ 2.89 & 38.39 $\pm$ 2.60 \\
        33 & \texttt{Qwen3-30b-a3b} & 54.77 $\pm$ 2.66 & 37.78 $\pm$ 3.32 \\
        34 & \texttt{GPT-$3.5$-turbo-0125} & 47.08 $\pm$ 2.11 & 36.17 $\pm$ 2.70 \\
        35 & \texttt{Reka-flash-3} & 41.52 $\pm$ 3.03 & 34.94 $\pm$ 3.02 \\
        36 & \texttt{DeepSeek-r1-distill-Qwen-32b} & 39.96 $\pm$ 3.25 & 31.05 $\pm$ 1.99 \\
        37 & \texttt{Granite-3.2-8b-it} & 30.00 $\pm$ 2.24 & \underline{30.86 $\pm$ 3.95} \\
        38 & \texttt{Deepthink-reasoning-14b} & 44.57 $\pm$ 2.45 & 30.45 $\pm$ 2.42 \\
        39 & \texttt{Qwen$2.5$-14b-it} & 43.46 $\pm$ 2.07 & 29.38 $\pm$ 2.79 \\
        40 & \hlcgreen{\texttt{Always C}} & - & 28.89 $\pm$ 2.56 \\\midrule\midrule
        41 & \texttt{QwQ-32b} & 51.36 $\pm$ 2.17 & 27.49 $\pm$ 2.56 \\
        42 & \hlcgreen{\texttt{Always B}} & - & 26.50 $\pm$ 3.24\\
        43 & \texttt{Deepseek-R1-distill-Qwen-14b} & 42.84 $\pm$ 2.64 & 25.31 $\pm$ 3.01 \\
        44 & \texttt{Deepthink-reasoning-7B} & 30.33 $\pm$ 2.35 & 25.14 $\pm$ 2.58 \\
        45 & \hlcgreen{\texttt{Random Sampling}} & - & 24.98 $\pm$ 2.16\\
        46 & \texttt{Qwen$2.5$-7B-it} & 29.84 $\pm$ 2.89 & 23.99 $\pm$ 1.99 \\
        47 & \hlcgreen{\texttt{Always D}} & - & 23.50 $\pm$ 2.91 \\
        48 & \texttt{s$1.1$-32b} & 47.37 $\pm$ 3.05 & 22.76 $\pm$ 1.60 \\
        49 & \hlcgreen{\texttt{Always A}} & - & 20.78 $\pm$ 1.75\\
        50 & \texttt{Gemini-$2.5$-pro} & 74.65 $\pm$ 2.36 & 14.06 $\pm$ 1.77 \\
        51 & \texttt{Grok-2-latest} & 59.26 $\pm$ 2.25 & 8.11 $\pm$ 2.14 \\
        52 & \texttt{Sonar-reasoning} & 57.84 $\pm$ 2.54 & 7.65 $\pm$ 2.20 \\
        53 & \texttt{Qwen$2.5$-14b} & 34.90 $\pm$ 1.90 & 7.61 $\pm$ 1.90 \\
        54 & \texttt{Llama-$3.3$-Nemotron-super-49B-v1} & 47.20 $\pm$ 3.04 & 3.09 $\pm$ 1.38 \\
        55 & \texttt{Llama-$3.3$-70b-it} & 61.48 $\pm$ 2.33 & 0.99 $\pm$ 0.71 \\
        56 & \texttt{French-Alpaca-Llama3-8b-it-v$1.0$} & 31.94 $\pm$ 2.88 & 0.08 $\pm$ 0.17 \\
        \bottomrule
   \end{tabular}%
   }
\end{table}

\section{Conclusion}
The automation of financial advice demands a transition from plausible generation to verifiable legal accuracy.
In this work, we introduced AEPC-QA, a gold-standard private benchmark derived from Quebec regulatory certification textbooks, and evaluated 51 LLMs to assess their readiness for this high-stakes domain.
Our results reveal a fundamental paradigm shift: performance is driven less by parametric memorization of the law and more by inference-time reasoning, with generalist reasoning models (e.g. \texttt{o3-2025-04-16}) significantly outperforming domain-specific French fine-tuned models.
While a RAG serves as a powerful equalizer for weaker models, we identified a critical \enquote{context distraction} vulnerability, where external retrieval paradoxically degrades the performance of certain high-capability models.
This instability, combined with the persistent gap between proprietary and open-weight architectures, suggests that autonomous deployment is premature.
Future research must focus on closing this gap by distilling reasoning and developing robustness interventions to mitigate context sensitivity.
Furthermore, evaluation protocols should evolve from multiple-choice selection to generative justification and multi-turn conversation, ensuring models cannot only select the correct advice but also explicitly cite the supporting regulatory frameworks.

\clearpage
\subsubsection*{Acknowledgments}
This research was made possible thanks to the support of a Canadian insurance company, NSERC research grant RDCPJ 537198-18 and FRQNT doctoral research grant. We thank the reviewers for their comments regarding our work.

\subsubsection*{Use of Large Language Models}
This work utilized Large Language Models (specifically Gemini) to assist with the linguistic refinement, typographical errors, and LaTeX formatting of the manuscript (table generation, figure improvements). 
The authors retained full control over the scientific content, data analysis, and conclusions. 
Additionally, as this paper presents a benchmark for LLMs, the outputs of various models (e.g. GPT-5, Claude, Qwen) were generated and analyzed as the primary subject of our research.

\subsubsection*{Ethical Considerations}
The deployment of LLMs in the financial sector extends beyond technical feasibility; it introduces profound ethical and legal challenges that demand rigorous scrutiny.
As our study operates at the intersection of consumer protection, civil liability, and automated decision-making, we identify three primary ethical dimensions inherent to this work.

\paragraph{Access vs. Safety}
The implementation of Bill 141 in Quebec has created a paradoxical \enquote{advice gap}: while digital distribution increases market efficiency, it removes the safety net of professional human oversight \citep{protectionepargnants}.
Our work aims to fill this void with automated QA systems. 
However, this creates an ethical tension.
While a 79\% accuracy rate (achieved by \texttt{o3-2025-04-16} using a RAG architecture) represents a significant technological milestone, the remaining 21\% error rate poses an acceptable risk in creative writing but unacceptable liability in financial advisory.
As illustrated by the \textit{Moffatt v. Air Canada} ruling \citep{air_canada}, automated agents are legally treated as extensions of the corporation.
Deploying models that demonstrate \enquote{Context Distraction}, where performance collapses under RAG, could constitute systemic negligent misrepresentation, potentially exposing vulnerable consumers to uninsured losses.
Therefore, we posit that \enquote{human-in-the-loop} architectures remain an ethical imperative until verifiable safety guarantees (e.g. conformal prediction sets) can be established.

\paragraph{Jurisdictional Alignment as an Ethical Constraint}
We highlight a critical, often overlooked ethical issue in Legal NLP: jurisdictional bias.
Most \enquote{legal} LLMs are trained on English data rooted in Common Law systems (e.g. US, UK).
Deploying such models in a Civil Law jurisdiction like Quebec is not merely a technical error; it is an ethical failure of the \enquote{legal}.
A model that hallucinates a \enquote{Common Law duty of care} or \enquote{punitive damages} (concepts often foreign or differently applied in Quebec Civil Law) creates a \enquote{legal mirage}, misleading users with plausible but jurisdictionally invalid advice.
Our decision to exclude English-centric legal models (e.g. \textit{SaulLM}) was driven by this ethical commitment to preventing cross-jurisdictional contamination, ensuring that the advice rendered respects the specific statutory framework of the Civil Code of Quebec.

\paragraph{Data Propriety and Reproducibility}
Finally, we address the tension between open science and intellectual property.
Academic rigour typically demands the public release of evaluation benchmarks.
However, the AEPC-QA corpus is derived from proprietary certification handbooks essential to the integrity of the regulatory exam process.
Releasing this data would not only violate copyright but could compromise the certification mechanism itself (by leaking potential exam content).
We resolved this dilemma by prioritizing data integrity and legal compliance over open release, opting instead to provide detailed methodology and synthetic examples (as seen in \autoref{fig:qnaprompt_exemple}).
This decision underscores a broader ethical reality in specialized domain research: high-value, \enquote{gold-standard} evaluation data will increasingly remain private, necessitating new frameworks for trusted, non-public benchmarking (e.g. code-to-data paradigms).

\bibliography{iclr2026_conference}

@article{Beauchemin2023RISC,
	author = {Beauchemin, David and Khoury, Richard},
	journal = {Proceedings of the Canadian Conference on Artificial Intelligence},
	year = {2023},
	month = {jun 5},
	note = {https://caiac.pubpub.org/pub/k18zu6c9},
	publisher = {Canadian Artificial Intelligence Association},
	title = {{RISC: Generating Realistic Synthetic Bilingual Insurance Contract}},
}

@inproceedings{beauchemin-etal-2020-generating,
    title = "Generating Intelligible Plumitifs Descriptions: Use Case Application with Ethical Considerations",
    author = "Beauchemin, David  and
      Garneau, Nicolas  and
      Gaumond, Eve  and
      D{\'e}ziel, Pierre-Luc  and
      Khoury, Richard  and
      Lamontagne, Luc",
    booktitle = "Proceedings of the International Conference on Natural Language Generation",
    month = dec,
    year = "2020",
    address = "Dublin, Ireland",
    publisher = "Association for Computational Linguistics",
    url = "https://aclanthology.org/2020.inlg-1.3",
    pages = "15--21",
}

@misc{amfbooks,
	title = {{Exam Preparation Manuals}},
	note = {Accessed online (30-06-2025) \url{https://manuels.lautorite.qc.ca/en/home.html}},
	year={2025},
	month = {June},
	author = {{Autorité des marchés financiers}},
}

@misc{Kincaid1975DerivationON,
  title={{Derivation of New Readability Formulas (Automated Readability Index, Fog Count and Flesch Reading Ease Formula) for Navy Enlisted Personnel}},
  author={J. Peter Kincaid and Robert P. Fishburne and Richard Lawrence Rogers and Brad S. Chissom},
  year={1975}
}

@article{article,
author = {Giampieri, Patrizia},
year = {2023},
month = {01},
pages = {119-137},
title = {{Is Machine Translation Reliable in the Legal Field? A Corpus-Based Critical Comparative Analysis for Teaching ESP at Tertiary Level}},
volume = {11},
journal = {ESP Today},
doi = {10.18485/esptoday.2023.11.1.6}
}

@misc{bill141,
        author                    =       {{National Assembly of Québec}},
        publisher                       =       {{National Assembly of Québec}},
        title                           =       {{An Act mainly to improve the regulation of the financial sector, the protection of deposits of money and the operation of financial institutions}},
        year                            =       2018,
        }

@inproceedings{sulem2018bleu,
      title={{BLEU is Not Suitable for the Evaluation of Text Simplification}}, 
      author={Elior Sulem and Omri Abend and Ari Rappoport},
      booktitle={Conference on Empirical Methods in Natural Language Processing},
      pages={738--744},
      year={2018},
      organization={Association for Computational Linguistics}
}

@article{wolf2020huggingfaces,
      title={{HuggingFace's Transformers: State-of-the-art Natural Language Processing}}, 
      author={Thomas Wolf and Lysandre Debut and Victor Sanh and Julien Chaumond and Clement Delangue and Anthony Moi and Pierric Cistac and Tim Rault and Rémi Louf and Morgan Funtowicz and Joe Davison and Sam Shleifer and Patrick von Platen and Clara Ma and Yacine Jernite and Julien Plu and Canwen Xu and Teven Le Scao and Sylvain Gugger and Mariama Drame and Quentin Lhoest and Alexander M. Rush},
      year={2020},
      journal={arXiv:1910.03771},
      primaryClass={cs.CL}
}

@misc{air_canada,
    author={Marisa Garcia}, 
    title="What Air Canada Lost In ‘Remarkable’ Lying AI Chatbot Case",
    year = 2024,
    url="https://www.forbes.com/sites/marisagarcia/2024/02/19/what-air-canada-lost-in-remarkable-lying-ai-chatbot-case/?sh=65b00f54696fl",
    note = {Accessed: 2024-03-10}
    }

@misc{air_Canada_court,
    author={{Moffatt \textit{v. Air Canada}}}, 
    note = "Civil Resolution Tribunal of British Columbia 149",
    year="2024"
    }

@inproceedings{louis2024interpretable,
  title={{Interpretable Long-Form Legal Question Answering With Retrieval-Augmented Large Language Models}},
  author={Louis, Antoine and van Dijck, Gijs and Spanakis, Gerasimos},
  booktitle={Proceedings of the AAAI Conference on Artificial Intelligence},
  volume={38:20},
  pages={22266--22275},
  year={2024}
}

@article{nuruzzaman2020intellibot,
  title={{IntelliBot: A Dialogue-Based Chatbot for the Insurance Industry}},
  author={Nuruzzaman, Mohammad and Hussain, Omar Khadeer},
  journal={Knowledge-Based Systems},
  volume={196},
  pages={105810},
  year={2020},
  publisher={Elsevier}
}

@article{li2022survey,
  title={{A Survey on Retrieval-Augmented Text Generation}},
  author={Li, Huayang and Su, Yixuan and Cai, Deng and Wang, Yan and Liu, Lemao},
  journal={arXiv:2202.01110},
  year={2022}
}

@book{protectionepargnants,
	author = {{Cinthia Duclos}},
	year = {2021},
	title = {La protection des épargnants dans l'industrie des services d'investissement: une analyse de l'influence des défaillances organisationnelles sous l'angle du swiss cheese model},
	isbn={9782897307486},
	publisher = {C{\'E}D{\'E}, Éditions Yvon Blais},
    pages={668}
}

@article{zheng2023judging,
  title={{Judging LLM-As-A-Judge With Mt-Bench and Chatbot Arena}},
  author={Zheng, Lianmin and Chiang, Wei-Lin and Sheng, Ying and Zhuang, Siyuan and Wu, Zhanghao and Zhuang, Yonghao and Lin, Zi and Li, Zhuohan and Li, Dacheng and Xing, Eric and others},
  journal={Advances in Neural Information Processing Systems},
  volume={36},
  pages={46595--46623},
  year={2023}
}

@article{gu2024survey,
  title={{A Survey on LLM-As-A-Judge}},
  author={Gu, Jiawei and Jiang, Xuhui and Shi, Zhichao and Tan, Hexiang and Zhai, Xuehao and Xu, Chengjin and Li, Wei and Shen, Yinghan and Ma, Shengjie and Liu, Honghao and others},
  journal={arXiv:2411.15594},
  year={2024}
}

@article{cao2025toward,
  title={{Toward Generalizable Evaluation in the LLM Era: A Survey Beyond Benchmarks}},
  author={Cao, Yixin and Hong, Shibo and Li, Xinze and Ying, Jiahao and Ma, Yubo and Liang, Haiyuan and Liu, Yantao and Yao, Zijun and Wang, Xiaozhi and Huang, Dan and others},
  journal={arXiv:2504.18838},
  year={2025}
}

@misc{azureaidoc,
	title = {{Azure AI Document Intelligence}},
	note = {Accessed online (15-02-2025) \url{https://azure.microsoft.com/en-us/products/ai-services/ai-document-intelligence}},
	year = {2025},
	author = {{Microsoft}},
}

@inproceedings{beauchemin2024quebec,
  title={{Quebec Automobile Insurance Question-Answering With Retrieval-Augmented Generation}},
  author={Beauchemin, David and Khoury, Richard and Gagnon, Zachary},
  booktitle={Proceedings of the Natural Legal Language Processing Workshop},
  pages={48--60},
  year={2024}
}

@article{ressel2024addressing,
  title={{Introduction to Common Crawl Datasets addressing the Notion of Trust Around ChatGPT in the High-Stakes Use Case of Insurance}},
  author={Ressel, Juliane and V{\"o}ller, Michaele and Murphy, Finbarr and Mullins, Martin},
  journal={Technology in Society},
  volume={78},
  pages={102644},
  year={2024},
  publisher={Elsevier}
}

@article{colombo2024saullm,
  title={{SauLLM-7B: A Pioneering Large Language Model for Law}},
  author={Colombo, Pierre and Pires, Telmo Pessoa and Boudiaf, Malik and Culver, Dominic and Melo, Rui and Corro, Caio and Martins, Andre FT and Esposito, Fabrizio and Raposo, Vera L{\'u}cia and Morgado, Sofia and others},
  journal={arXiv:2403.03883},
  year={2024}
}

@article{martinez2022poor,
  title={{Poor Writing, Not Specialized Concepts, Drives Processing Difficulty in Legal Language}},
  author={Mart{\'\i}nez, Eric and Mollica, Francis and Gibson, Edward},
  journal={Cognition},
  volume={224},
  pages={105070},
  year={2022},
  publisher={Elsevier}
}

@book{duclos2024droit,
  title={Droit des services d'investissement : Encadrement des intermédiaires financiers et protection des épargnants},
  author={Duclos, Cinthia and Cr{\^e}te, Raymonde and C{\^o}t{\'e}, Martin},
  year={2024},
  publisher={{\'E}ditions Yvon Blais},
  address={Montr{\'e}al},
  month={dec},
  note={Avec la collaboration de Salomé Paradis}
}

@inproceedings{roberts2020much,
  title={{How Much Knowledge Can You Pack Into the Parameters of a Language Model?}},
  author={Roberts, Adam and Raffel, Colin and Shazeer, Noam},
  booktitle={Proceedings of the Conference on Empirical Methods in Natural Language Processing},
  year={2020},
  organization={Association for Computational Linguistics}
}

@misc{open-llm-leaderboard-v2,
  author = {Clémentine Fourrier and Nathan Habib and Alina Lozovskaya and Konrad Szafer and Thomas Wolf},
  title = {{Open LLM Leaderboard v2}},
  year = {2024},
  publisher = {Hugging Face},
  howpublished = "\url{https://huggingface.co/spaces/open-llm-leaderboard/open_llm_leaderboard}",
}

@misc{deepseekai2025deepseekr1incentivizingreasoningcapability,
      title={{DeepSeek-R1: Incentivizing Reasoning Capability in LLMs via Reinforcement Learning}}, 
      author={DeepSeek-AI},
      year={2025},
      eprint={2501.12948},
      archivePrefix={arXiv},
      primaryClass={cs.CL},
      url={https://arxiv.org/abs/2501.12948}, 
}

@misc{alpaca,
    title={{French-Alpaca-Llama3-8B-Instruct-v1.0}},
    author={Pacifico, Jonathan},
    year={2024},
    url={https://huggingface.co/jpacifico/French-Alpaca-Llama3-8B-Instruct-v1.0}
}

@article{grattafiori2024llama,
  title={{The Llama 3 Herd of Models}},
  author={Grattafiori, Aaron and Dubey, Abhimanyu and Jauhri, Abhinav and Pandey, Abhinav and Kadian, Abhishek and Al-Dahle, Ahmad and Letman, Aiesha and Mathur, Akhil and Schelten, Alan and Vaughan, Alex and others},
  journal={arXiv:2407.21783},
  year={2024}
}

@misc{deepthink1,
    title={{Deepthink-Reasoning-7B}},
    author={Sakthi, Prithiv},
    year={2025},
    url={https://huggingface.co/prithivMLmods/Deepthink-Reasoning-7B}
}

@misc{deepthink2,
    title={{Deepthink-Reasoning-14B}},
    author={Sakthi, Prithiv},
    year={2025},
    url={https://huggingface.co/prithivMLmods/Deepthink-Reasoning-14B}
}

@article{hui2024qwen2,
  title={{Qwen2.5 Technical Report}},
  author={Hui, Binyuan and Yang, Jian and Cui, Zeyu and Yang, Jiaxi and Liu, Dayiheng and Zhang, Lei and Liu, Tianyu and Zhang, Jiajun and Yu, Bowen and Dang, Kai and others},
  journal={CoRR},
  year={2024}
}

@misc{reka,
    title={{Reka-flash-3}},
    author={{Reka AI}},
    year={2025},
    url={https://huggingface.co/RekaAI/reka-flash-3}
}

@misc{s11,
    title={{s1.1-32B}},
    author={{Simple Scaling}},
    year={2025},
    url={https://huggingface.co/simplescaling/s1.1-32B}
}

@article{rastogi2025magistral,
  title={Magistral},
  author={Rastogi, Abhinav and Jiang, Albert Q and Lo, Andy and Berrada, Gabrielle and Lample, Guillaume and Rute, Jason and Barmentlo, Joep and Yadav, Karmesh and Khandelwal, Kartik and Chandu, Khyathi Raghavi and others},
  journal={arXiv:2506.10910},
  year={2025}
}

@article{abdin2024phi,
  title={{Phi-4 Technical Report}},
  author={Abdin, Marah and Aneja, Jyoti and Behl, Harkirat and Bubeck, S{\'e}bastien and Eldan, Ronen and Gunasekar, Suriya and Harrison, Michael and Hewett, Russell J and Javaheripi, Mojan and Kauffmann, Piero and others},
  journal={arXiv:2412.08905},
  year={2024}
}

@misc{qwen3technicalreport,
      title={{Qwen3 Technical Report}}, 
      author={{Qwen Team}},
      year={2025},
      eprint={2505.09388},
      archivePrefix={arXiv},
      primaryClass={cs.CL},
      url={https://arxiv.org/abs/2505.09388}, 
}

@article{granite2024granite,
  title={{Granite 3.0 Language Models}},
  author={Granite Team, IBM},
  journal={URL: https://github. com/ibm-granite/granite-3.0-language-models},
  year={2024}
}

@article{nguyen2023brief,
  title={{A brief report on LawGPT 1.0: A virtual legal assistant based on GPT-3}},
  author={Nguyen, Ha-Thanh},
  journal={arXiv:2302.05729},
  year={2023}
}

@article{magesh2024hallucination,
  title={{Hallucination-Free? Assessing the Reliability of Leading AI Legal Research Tools}},
  author={Magesh, Varun and Surani, Faiz and Dahl, Matthew and Suzgun, Mirac and Manning, Christopher D and Ho, Daniel E},
  journal={Journal of Empirical Legal Studies},
  volume={22},
  number={2},
  pages={216--242},
  year={2025},
  publisher={Wiley Online Library}
}

@article{guha2023legalbench,
  title={{LegalBench: A Collaboratively Built Benchmark for Measuring Legal Reasoning in Large Language Models}},
  author={Guha, Neel and Nyarko, Julian and Ho, Daniel and R{\'e}, Christopher and Chilton, Adam and Chohlas-Wood, Alex and Peters, Austin and Waldon, Brandon and Rockmore, Daniel and Zambrano, Diego and others},
  journal={Advances in neural information processing systems},
  volume={36},
  pages={44123--44279},
  year={2023}
}

@inproceedings{fei2024lawbench,
  title={{Lawbench: Benchmarking Legal Knowledge of Large Language Models}},
  author={Fei, Zhiwei and Shen, Xiaoyu and Zhu, Dawei and Zhou, Fengzhe and Han, Zhuo and Huang, Alan and Zhang, Songyang and Chen, Kai and Yin, Zhixin and Shen, Zongwen and others},
  booktitle={Proceedings of the conference on empirical methods in natural language processing},
  pages={7933--7962},
  year={2024}
}

@inproceedings{hu2025fintrust,
  title={{Fintrust: A Comprehensive Benchmark of Trustworthiness Evaluation in Finance Domain}},
  author={Hu, Tiansheng and Hu, Tongyan and Bai, Liuyang and Zhao, Yilun and Cohan, Arman and Zhao, Chen},
  booktitle={Proceedings of the Conference on Empirical Methods in Natural Language Processing},
  pages={10110--10139},
  year={2025}
}

@article{ho2024hallucinating,
  title={{Hallucinating Law: Legal Mistakes with Large Language Models are Pervasive}},
  author={Dahl, Matthew and Magesh, Varun and Suzgun, Mirac and Ho, Daniel E},
  journal={Law, regulation, and policy},
  year={2024}
}

@article{Barua2025,
  author  = {Barua, Josh and Eisape, Seun and Yin, Kayo and Suhr, Alane},
  title   = {{Long Chain-of-Thought Reasoning Across Languages}},
  journal = {arXiv:2508.14828},
  year    = {2025},
  url     = {https://arxiv.org/abs/2508.14828}
}

@article{newcombe1998two,
title   = {{Two-Sided Confidence Intervals for the Single Proportion: Comparison of Seven Methods}},
author  = {Newcombe, Robert G},
journal = {Statistics in Medicine},
volume  = {17},
number  = {8},
pages   = {857--872},
year    = {1998}
}

@misc{bercovich2025llamanemotronefficientreasoningmodels,
        title={{Llama-Nemotron: Efficient Reasoning Models}},         
        author={Akhiad Bercovich and Itay Levy and Izik Golan and Mohammad Dabbah and Ran El-Yaniv and Omri Puny and Ido Galil and Zach Moshe and Tomer Ronen and Najeeb Nabwani and  Ido Shahaf and Oren Tropp and Ehud Karpas and Ran Zilberstein and Jiaqi Zeng and Soumye Singhal and Alexander Bukharin and Yian Zhang and Tugrul Konuk and Gerald Shen and Ameya Sunil Mahabaleshwarkar and Bilal Kartal and Yoshi Suhara and Olivier Delalleau and Zijia Chen and Zhilin Wang and David Mosallanezhad and Adi Renduchintala and Haifeng Qian and Dima Rekesh and Fei Jia and Somshubra Majumdar and Vahid Noroozi and Wasi Uddin Ahmad and Sean Narenthiran and Aleksander Ficek and Mehrzad Samadi and Jocelyn Huang and Siddhartha Jain and Igor Gitman and Ivan Moshkov and Wei Du and Shubham Toshniwal and George Armstrong and Branislav Kisacanin and Matvei Novikov and Daria Gitman and Evelina Bakhturina and Jane Polak Scowcroft and John Kamalu and Dan Su and Kezhi Kong and Markus Kliegl and Rabeeh Karimi and Ying Lin and Sanjeev Satheesh and Jupinder Parmar and Pritam Gundecha and Brandon Norick and Joseph Jennings and Shrimai Prabhumoye and Syeda Nahida Akter and Mostofa Patwary and Abhinav Khattar and Deepak Narayanan and Roger Waleffe and Jimmy Zhang and Bor-Yiing Su and Guyue Huang and Terry Kong and Parth Chadha and Sahil Jain and Christine Harvey and Elad Segal and Jining Huang and Sergey Kashirsky and Robert McQueen and Izzy Putterman and George Lam and Arun Venkatesan and Sherry Wu and Vinh Nguyen and Manoj Kilaru and Andrew Wang and Anna Warno and Abhilash Somasamudramath and Sandip Bhaskar and Maka Dong and Nave Assaf and Shahar Mor and Omer Ullman Argov and Scot Junkin and Oleksandr Romanenko and Pedro Larroy and Monika Katariya and Marco Rovinelli and Viji Balas and Nicholas Edelman and Anahita Bhiwandiwalla and Muthu Subramaniam and Smita Ithape and Karthik Ramamoorthy and Yuting Wu and Suguna Varshini Velury and Omri Almog and Joyjit Daw and Denys Fridman and Erick Galinkin and Michael Evans and Katherine Luna and Leon Derczynski and Nikki Pope and Eileen Long and Seth Schneider and Guillermo Siman and Tomasz Grzegorzek and Pablo Ribalta and Joey Conway and Trisha Saar and Ann Guan and Krzysztof Pawelec and Shyamala Prayaga and Oleksii Kuchaiev and Boris Ginsburg and Oluwatobi Olabiyi and Kari Briski and Jonathan Cohen and Bryan Catanzaro and Jonah Alben and Yonatan Geifman and Eric Chung and Chris Alexiuk},
        year={2025},
        eprint={2505.00949},
        archivePrefix={arXiv},
        primaryClass={cs.CL},
        url={https://arxiv.org/abs/2505.00949},
  }
\bibliographystyle{iclr2026_conference}

\appendix
\section{Selected LLM Details}
\label{an:selectedllmdetails}
We present in \autoref{tab:selectedllm} the comprehensive suite of open-source LLMs we could fit on our hardware (see \autoref{an:hardware}), detailing their origins and respective sizes, while in \autoref{tab:selectedprivatellm}, we present the comprehensive suite of private LLMs benchmarked in our study.
The selection was curated to cover a wide spectrum of parameter counts, and to include those with specializations in French ($\Upsilon$) or reasoning ($\Gamma$).
All LLMs are downloaded from the \href{https://huggingface.co/models}{HuggingFace Model repository} \citep{wolf2020huggingfaces} using default parameters.

\begin{table*}[ht!]
    \centering
    \caption{The selected open-source LLMs used in our work (16 models), along with their source and size. \enquote{$\Upsilon$} are models that have a specialization in French, while \enquote{$\Gamma$} are models marketed as reasoning LLMs.}
    \label{tab:selectedllm}
    \resizebox{\textwidth}{!}{%
    \begin{tabular}{llcllc}
        \toprule
        \textbf{LLM} & \textbf{Source} & \textbf{Size} & \textbf{LLM} & \textbf{Source} & \textbf{Size} \\
        \midrule
        \texttt{DeepSeek-R1-distill-Qwen-14b} ($\Gamma$) & \citet{deepseekai2025deepseekr1incentivizingreasoningcapability} & 14.8B & \texttt{Phi-4} & \citet{abdin2024phi} & 14.7B \\
        \texttt{DeepSeek-R1-distill-Qwen-32b} ($\Gamma$) & \citet{deepseekai2025deepseekr1incentivizingreasoningcapability} & 32.8B & \texttt{Qwen2.5-14b} & \citet{hui2024qwen2} & 14.7B \\
        \texttt{Deepthink-reasoning-14b} ($\Gamma$) & \citet{deepthink2} & 14.8B & \texttt{Qwen2.5-14b-it} & \citet{hui2024qwen2} & 14.7B \\
        \texttt{Deepthink-reasoning-7b} ($\Gamma$) & \citet{deepthink1} & 7.62B & \texttt{Qwen2.5-7b-it} & \citet{hui2024qwen2} & 7.6B \\
        \texttt{French-Alpaca-Llama3-8b-it} ($\Upsilon$, $\Gamma$) & \citet{alpaca} & 8.03B & \texttt{Qwen3-30b-a3b} & \citet{qwen3technicalreport} & 30B \\
        \texttt{Granite-3.2-8b-it} & \citet{granite2024granite} & 8.17B & \texttt{QwQ-32b} ($\Gamma$) & \citet{qwen3technicalreport} & 32.8B \\
        \texttt{Llama-3.3-70b-it} & \citet{grattafiori2024llama} & 70B & \texttt{Reka-flash-3} ($\Gamma$) & \citet{reka} & 21B \\
        \texttt{Llama-3.3-Nemotron-49B} & \citet{bercovich2025llamanemotronefficientreasoningmodels} & 49B & \texttt{S1.1-32b} ($\Gamma$) & \citet{s11} & 32.8B \\
        \bottomrule
    \end{tabular}
    }
\end{table*}

\begin{table*}[ht!]
    \centering
    \caption{The selected private LLMs used in our work, along with their source. \enquote{$\Gamma$} indicates models marketed as reasoning LLMs.}
    \label{tab:selectedprivatellm}
    \resizebox{\textwidth}{!}{%
    \begin{tabular}{llll}
        \toprule
        \textbf{LLM} & \textbf{Source} & \textbf{LLM} & \textbf{Source} \\
        \cmidrule{1-2} \cmidrule{3-4}
        \texttt{Claude-3.5-Haiku-20241022} & Anthropic & \texttt{GPT-4o-mini-2024-07-18} & OpenAI \\
        \texttt{Claude-3.7-Sonnet-20250219} ($\Gamma$) & Anthropic & \texttt{Grok-2-latest} & xAI \\
        \texttt{Claude-Opus-4-20250514} ($\Gamma$) & Anthropic & \texttt{Grok-3-fast-latest} ($\Gamma$) & xAI \\
        \texttt{Claude-Sonnet-4-20250514} ($\Gamma$) & Anthropic & \texttt{Grok-3-latest} ($\Gamma$) & xAI \\
        \texttt{DeepSeek-chat} & DeepSeek & \texttt{Grok-3-mini-fast-latest} ($\Gamma$) & xAI \\
        \texttt{DeepSeek-reasoner} ($\Gamma$) & DeepSeek & \texttt{Grok-3-mini-latest} ($\Gamma$) & xAI \\
        \texttt{Gemini-2.5-flash} & Google & \texttt{Magistral-medium-2506} & \citet{rastogi2025magistral} \\
        \texttt{Gemini-2.5-pro} ($\Gamma$) & Google & \texttt{Mistral-large-latest} ($\Gamma$) & Mistral \\
        \texttt{GPT-3.5-turbo-0125} & OpenAI & \texttt{Mistral-medium-2505} & Mistral \\
        \texttt{GPT-4-0613} & OpenAI & \texttt{o1-2024-12-17} ($\Gamma$) & OpenAI \\
        \texttt{GPT-4.1-2025-04-14} & OpenAI & \texttt{o1-mini-2024-09-12} ($\Gamma$) & OpenAI \\
        \texttt{GPT-4.1-mini-2025-04-14} & OpenAI & \texttt{o3-2025-04-16} ($\Gamma$) & OpenAI \\
        \texttt{GPT-4.1-nano-2025-04-14} & OpenAI & \texttt{o3-mini-2025-01-31} ($\Gamma$) & OpenAI \\
        \texttt{GPT-4.5-preview-2025-02-27} & OpenAI & \texttt{o4-mini-2025-04-16} ($\Gamma$) & OpenAI \\
        \texttt{Pixtral-large-latest} & Mistral & \texttt{Qwen-max} ($\Gamma$) & Qwen \\
        \texttt{R1-1776} ($\Gamma$) & Private & \texttt{Sonar} & Perplexity \\
        \texttt{Sonar-deep-research} & Perplexity & \texttt{Sonar-reasoning} & Perplexity \\
        \texttt{Sonar-reasoning-pro} & Perplexity & & \\
        \bottomrule
    \end{tabular}
    }
\end{table*}

\section{Hardware and Private LLM Inference Budget}
\label{an:hardware}
\subsection{Hardware}
We rely on three NVIDIA RTX 6000 ADA with 49 GB of memory, without memory pooling; thus, the maximum size we can fit is around 32B parameters to achieve a sufficient batch size to process the experiment in a reasonable timeframe (i.e. a month or so).

\subsection{Private LLM Inference Budget}
We allocated approximately 2,000 USD for using private LLM APIs (e.g. OpenAI, Anthropic) during development, prototyping, and prompt tuning. For the complete inference loop across all selected private LLMs and tasks, we allocated a budget of nearly \$ 7,500 USD.
It took approximately four weeks to process all private model inference in parallel.

\end{document}